\documentclass{article}
\usepackage{ijcai23}

\usepackage{times}
\usepackage{soul}
\usepackage{url}
\usepackage[utf8]{inputenc}
\usepackage[small]{caption}
\usepackage{graphicx}
\usepackage{amsmath}
\usepackage{amsthm}
\usepackage{booktabs}
\usepackage{algorithm}
\usepackage{algorithmic}
\usepackage[switch]{lineno}

\usepackage{xcolor,pifont}
\usepackage{multirow,makecell}
\usepackage{bm}
\usepackage{subfigure}
\usepackage{amssymb}

\title{Hawkes Process Based on Controlled Differential Equations}

\author{
Minju Jo$^{}$\thanks{These authors contributed equally.}\and
Seungji Kook$^{*}$\And
Noseong Park\\
\affiliations
Yonsei University, Seoul, South Korea\\
\emails
\{alflsowl12,202132139,noseong\}@yonsei.ac.kr
}

\begin{document}

\maketitle

\begin{abstract}
    Hawkes processes are a popular framework to model the occurrence of sequential events, i.e., occurrence dynamics, in several fields such as social diffusion. In real-world scenarios, the inter-arrival time among events is \emph{irregular}. However, existing neural network-based Hawkes process models not only i) fail to capture such complicated irregular dynamics but also ii) resort to heuristics to calculate the log-likelihood of events since they are mostly based on neural networks designed for regular discrete inputs. To this end, we present the concept of Hawkes process based on controlled differential equations (HP-CDE), by adopting the neural controlled differential equation (neural CDE) technology which is an analogue to \emph{continuous} RNNs. Since HP-CDE continuously reads data, i) irregular time-series datasets can be properly treated preserving their uneven temporal spaces, and ii) the log-likelihood can be exactly computed. Moreover, as both Hawkes processes and neural CDEs are first developed to model complicated human behavioral dynamics, neural CDE-based Hawkes processes are successful in modeling such occurrence dynamics. In our experiments with 4 real-world datasets, our method outperforms existing methods by non-trivial margins.
\end{abstract}

\section{Introduction}

Real-world phenomena typically correspond to the occurrence of sequential events with \emph{irregular} time intervals and \emph{numerous} event types, ranging from online social network activities to personalized healthcare and so on \cite{10.1145/2783258.2783401,enguehard2020neural,stoyan2000recent,mohler2011self,ogata1999seismicity}.
Hawkes processes and Poisson point process are typically used to model those sequential events~\cite{10.2307/2334319,miles1970homogeneous,streit2010poisson}. However, their basic assumptions are too stringent to model such complicated dynamics, e.g., all past events should influence the occurrence of the current event. To this end, many advanced techniques have been proposed for the past several years, ranging from classical recurrent neural network (RNN) based models such as RMTPP~\cite{10.1145/2939672.2939875} and NHP~\cite{mei2017neuralhawkes} to recent transformer models like SAHP~\cite{pmlr-v119-zhang20q} and THP~\cite{10.5555/3524938.3526022}. Even so, they still do not treat data in a fully continuous way but resort to heuristics, which is sub-optimal in processing irregular events~\cite{NIPS2018_7892,choi2021ltocf,NIPS2019_9497}. Likewise, their heuristic approaches to model the continuous time domain impede solving the multivariate integral of the log-likelihood calculation in Eq.~\eqref{eq:logProb}, leading to approximation methods such as the Monte Carlo sampling (cf. Table~\ref{tbl:comparison}). As a consequence, the strict constraint and/or the inexact calculation of the log-likelihood may induce inaccurate predictions.

\begin{table}[t]
\small
\setlength{\tabcolsep}{2pt}
  \begin{center}

    \label{tab:table1}
    \begin{tabular}{c|c|c}
    \Xhline{3\arrayrulewidth}
      \textbf{Model} & \textbf{Exact log-likelihood} & \textbf{How to model dynamics}\\
      \Xhline{2\arrayrulewidth}
      NHP, SAHP,    & \multirow{2}{*}{X} & \multirow{2}{*}{Discrete}\\
      THP     &  & \\
      \Xhline{2\arrayrulewidth}
      HP-CDE  & \begin{tabular}[c]{@{}c@{}}O\\($\lambda^*$ is continuous.)\end{tabular} & \begin{tabular}[c]{@{}c@{}}Continuous \&\\ robust to irregular dynamics\end{tabular}\\
      \Xhline{3\arrayrulewidth}
    \end{tabular}
        \caption{Comparison of neural network-based Hawkes process models. $\lambda^*$ denotes the conditional intensity function (cf. Eqs.~\eqref{eq:logProb},~\eqref{eq:exact}, and~\eqref{eqn:lambda_prediction}).}\label{tbl:comparison}
  \end{center}
\end{table}

In this work, therefore, we model the occurrence dynamics based on differential equations, not only directly handling the sequential events in a continuous time domain but also exactly solving the integral of the log-likelihood. One more inspiration of using differential equations is that they have shown several non-trivial successes in modeling human behavioral dynamics~\cite{poli2019graph,NIPS2019_8773,lightmove} --- in particular, we are interested in controlled differential equations. To our knowledge, therefore, we first answer the question of whether occurrence dynamics can be modeled as controlled differential equations.

Controlled differential equations (CDEs~\cite{lyons2004differential}) are one of the most suitable ones for building human behavioral models. CDEs were first developed by a financial mathematician to model complicated dynamics in financial markets which is a typical application domain of Hawkes processes since financial transactions are temporal point processes. In particular, neural controlled differential equations (neural CDEs~\cite{kidger2020neural}), whose initial value problem (IVP) is written as below, are a set of techniques to learn CDEs from data with neural networks:
\begin{linenomath}\begin{align}\begin{split}
\mathbf{h}(t_b) 
&= \mathbf{h}(t_a) + \int_{t_a}^{t_b} f(\mathbf{h}(t);\mathbf{\theta}_f) dZ(t)\\
&= \mathbf{h}(t_a) + \int_{t_a}^{t_b} f(\mathbf{h}(t);\mathbf{\theta}_f) \frac{dZ(t)}{dt} dt,\label{eq:ncde2}
\end{split}\end{align}\end{linenomath}
where $f$ is a CDE function, and $\mathbf{h}(t)$ is a hidden vector at time $t$. $Z(t)$ is a continuous path created from discrete sequential observations (or events) $\{(\mathbf{z}_j, t_j)\}_{j=a}^b$ by an appropriate algorithm\footnote{One can use interpolation algorithms or neural networks for creating $Z(t)$ from $\{(\mathbf{z}_j, t_j)\}_{j=a}^b$ ~\cite{kidger2020neural}.}, where in our case, $\mathbf{z}_j$ is a vector containing the information of $j$-th occurrence, and $t_j \in [t_a,t_b]$ contains the time-point of the occurrence, i.e., $t_j < t_{j+1}$. Note that neural CDEs keep reading the time-derivative of $Z(t)$ over time, denoted $\dot{Z}(t) := \frac{dZ(t)}{dt}$, and for this reason, neural CDEs are in general, considered as \emph{continuous} RNNs. In addition, NCDEs are known to be superior in processing irregular time series~\cite{lyons2004differential}.

Given the neural CDE framework, we propose \textbf{\underline{H}}awkes \textbf{\underline{P}}rocess based on \textbf{\underline{C}}ontrolled \textbf{\underline{D}}ifferential \textbf{\underline{E}}quations (HP-CDE). We let $\mathbf{z}_j$ be the sum of the event embedding and the positional embedding and create a path $Z(t)$ with the linear interpolation method which is a widely used interpolation algorithm for neural CDEs (cf. Figure~\ref{fig:onecol2}). To get the exact log-likelihood, we use an ODE solver to calculate the non-event log-likelihood. Calculating the non-event log-likelihood involves the integral problem in Eq.~\eqref{eq:logProb}, and our method can solve it exactly since conditional intensity function $\lambda^*$, which indicates an instantaneous probability of an event, is defined in a continuous manner over time by the neural CDE technology. In addition, we have three prediction layers to predict the event log-likelihood, the event type, and the event occurrence time (cf. Eqs.~\eqref{eq:newLogProb}, ~\eqref{eqn:type prediction}, ~\eqref{eqn:time prediction} and Figure~\ref{fig:onecol3}).

We conduct event prediction experiments with 4 datasets and 4 baselines. Our method shows outstanding performance in all three aspects: i) event type prediction, ii) event time prediction, and iii) log-likelihood. Our contributions are as follows:
\begin{enumerate}
    \item We model the \emph{continuous} occurrence dynamics under the framework of neural CDE whose original theory was developed for describing \emph{irregular non-linear} dynamics. Many real-world Hawkes process datasets have irregular inter-arrival times of events.
    \item We then exactly solve the integral problem in Eq.~\eqref{eq:logProb} to calculate the non-event log-likelihood, which had been done typically through heuristic methods before our work.
\end{enumerate}

\section{Preliminaries}

\subsection{Multivariate Point Processes}

Multivariate point processes are a generative model of an event sequence $X = \{(k_j, t_j)\}_{j=1}^N$ and $x_j = (k_j, t_j)$ indicates $j$-th event in the sequence. This event sequence is a subset of an event stream under a continuous time interval $[t_1,t_N]$, and an observation $x_j$ at time $t_j$ has an event type $k_j \in \{1,\cdots,K\}$,  where $K$ is total number of event types. The arrival time of events is defined as $t_1<t_2<\cdots<t_N$. The point process model learns a probability for every $(k, t)$ pair, where $k \in \{1,\cdots,K\}, \ t \in [t_1,t_N]$. 

The key feature of multivariate point processes is the intensity function $\lambda_k(t)$, i.e., the probability that a type-$k$ event occurs at the infinitesimal time interval $[t,t+dt)$. The Hawkes process, one popular point process model, assumes that the intensity $\lambda_k(t)$ of type $k$ can be calculated by past events before $t$, so-called history $\mathcal{H}_t$, and its form is as follows:
\begin{linenomath}\begin{align}\label{eq:hawkesIntensity}
\lambda^*_k(t) := \lambda_k(t | \mathcal{H}_t) =  \mu_k  +   \sum_{j:t_j<t} \psi_k(t -  t_j),\end{align}\end{linenomath}
where $\lambda^*(t)=\sum_{k=1}^K\lambda^*_k(t)$, $\mu_k$ is the base intensity, and $\psi_k(\cdot)$ is a pre-determined decaying function for type $k$. We use $*$ to represent conditioning on the history $\mathcal{H}_t$. According to the formula, all the past events affect the probability of new event occurrence with different influences. However, the intensity converges to the base intensity if the decaying function becomes close to zero. 

Currently, a deep learning mechanism is applied to Hawkes processes by parameterizing the intensity function. For instance, RNNs are used in the neural Hawkes process (NHP)~\cite{mei2017neuralhawkes}, and its intensity function is defined as follows:
\begin{linenomath}\begin{align}\label{eq:neuralHawkesIntensity}
\lambda^*(t) =  \sum_{k=1}^K \phi_k(\mathbf{w}_k^\top\mathbf{h}(t)), \quad t \in [t_1,t_N],\end{align}\end{linenomath}
where $\phi_k(\cdot)$ is the softplus function, $\mathbf{h}(t)$ is a hidden state from RNNs, and $\mathbf{w}_k$ is a weight for each event type. The softplus function keeps intensity values positive. However, one downside of NHP is that RNN-based models assume that events have regular intervals. Thus, one of the main issues in NHP is how to fit a model to a continuous irregular time domain.

\subsection{Neural Network-based Hawkes Processes}\label{sec:hp_history}

Hawkes processes are a popular temporal predicting framework in various fields since it predicts both \emph{when}, which \emph{type} of events would happen with mathematical approaches. It is especially widely used in sociology fields to capture the diffusion of information~\cite{hardiman2013critical,kobayashi2016tideh,da2014hawkes}, seismology fields to model when earthquakes and aftershocks occur, medical fields to track the status of patients~\cite{choi2015constructing,garetto2021time}, and so on.

For enhancing the performance of Hawkes processes, a lot of deep learning approaches have been applied. The two basic approaches are the recurrent marked temporal point process (RMTPP~\cite{10.1145/2939672.2939875}) and the neural Hawkes process (NHP~\cite{mei2017neuralhawkes}). RMTPP is the first model that combines RNNs into point processes, and NHP is a Hawkes process model with an RNN-parameterized intensity function. Based on NHP, the self-attentive Hawkes process (SAHP~\cite{pmlr-v119-zhang20q}) attaches self-attention modules to reflect the relationships between events. Additionally, the transformer Hawkes process (THP~\cite{10.5555/3524938.3526022}) uses the transformer technology~\cite{NIPS2017_3f5ee243}, one of the most popular structures in natural language processing, to capture both short-term and long-term temporal dependencies of event sequences.

One important issue of neural network-based Hawkes process is how to handle irregular time-series datasets. To deal with this issue, NHP uses continuous-time LSTMs, whose memory cell exponentially decays. SAHP and THP both employ modified positional encoding schemes to represent irregular time intervals since the conventional encoding assumes regular spaces between events. However, all mentioned approaches still do not explicitly process irregular time-series. In contrast to them, our HP-CDE is robust to irregular time-series since the original motivation of neural CDEs is better processing irregular time-series by constructing continuous RNNs.

\subsection{Neural Controlled Differential Equations as continuous RNNs} 
\begin{figure}[t]
    \centering
    \subfigure{\includegraphics[width=0.9\columnwidth]{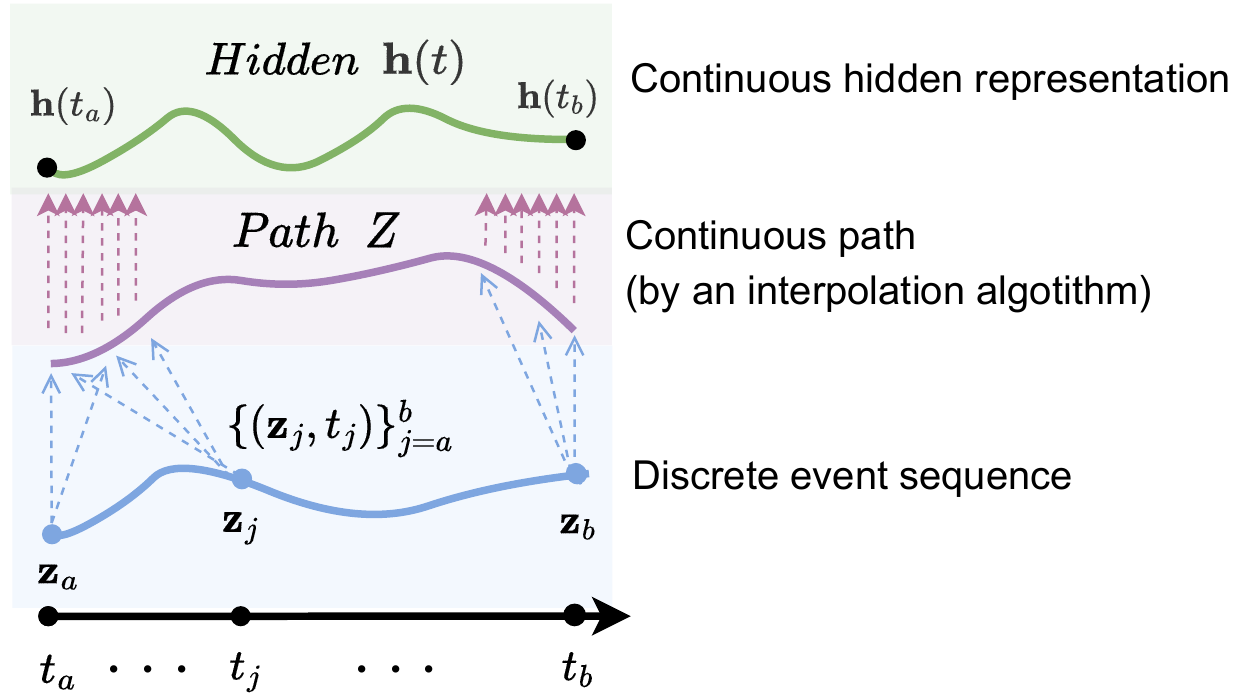}}
    \caption{Visualization of the continuous hidden state of the neural CDE model}
    \label{fig:cde}
\end{figure}

Neural controlled differential equations (neural CDEs) are normally regarded as a continuous analogue to RNNs since they process the time-derivative of the continuous path $Z(t)$. Especially, neural CDEs retain their continuous properties by using the interpolated path $Z$ made of discrete data $\{(\mathbf{z}_j, t_j)\}^b_{j=a}$ and solving the Riemann-Stieltjes integral to get $\mathbf{h}(t_b)$ from $\mathbf{h}(t_a)$ as shown in Eq.~\eqref{eq:ncde2} --- in particular, this problem to derive $\mathbf{h}(t_b)$ from the initial condition $\mathbf{h}(t_a)$ is known as initial value problem (IVP) (cf. Figure~\ref{fig:cde}). At first, to make the interpolated continuous path $Z$, linear interpolation or natural cubic spline interpolation is generally used among several interpolation methods. Then, we use existing ODE solvers to solve the Riemann-Stieltjes integral problem with $\dot{\mathbf{h}}(t) := \frac{d\mathbf{h}(t)}{dt}= f(\mathbf{h}(t);\mathbf{\theta}_f) \frac{dZ(t)}{dt}$.

\subsection{Maximum Likelihood Estimation in Temporal Point Process}
Most of the neural temporal point process frameworks choose the maximum likelihood estimation (MLE)~\cite{10.1214/aoms/1177706538} as one of the main training objectives. In order to enable the MLE training, getting the log-probability of every sequence $X$ is required, which consists of formulas using intensity functions conditioned on the history $\mathcal{H}_t$=$\{(k_j, t_j):t_j<t\}$. Thus, log-probability for any event sequence $X$ whose events are observed in an interval $[t_1, t_N]$ is as follows:
\begin{linenomath}\begin{align}\label{eq:logProb}
\log p(X) = \sum_{j=1}^{N} \log \lambda^*(t_j) - \int^{t_N}_{t_1} \lambda^*(t)dt,
\end{align}\end{linenomath}where $\sum_{j=1}^{N} \log \lambda^*(t_j)$ denotes the event log-likelihood and $\int^{t_N}_{t_1} \lambda^*(t)dt$ means the non-event log-likelihood. Non-event log-likelihood represents sum of the infinite number of non-events' log-probabilities in $[t_1,t_N]$, except the infinitesimal times when the event occurs. In the case of the event log-likelihood, it is comparably easy to compute as the formula is simply a sum of the intensity functions. However, it is challenging to compute the non-event log-likelihood, due to its integral computation. 
Due to the difficulty, NHP, SAHP, THP and many other models use approximation methods, such as Monte Carlo integration~\cite{10.5555/1051451} and numerical integration methods~\cite{nu1}, to get the value. However, since those methods do not exactly solve the integral problem, numerical errors are inevitable.

\section{Proposed Method}

In this section, we describe our \emph{explicitly continuous} Hawkes process model, called HP-CDE, based on the neural CDE framework which is considered as continuous RNNs. Owing to the continuous property of the proposed model, the exact log-likelihood, especially for the non-event log-likelihood part with its challenging integral calculation, can also be computed through ODE solvers. That is, our proposed model reads event sequences with irregular inter-arrival times in a continuous manner, and exactly computes the log-likelihood.

\subsection{Overall Workflow}

\begin{figure}[t] 
\begin{center}
\includegraphics[width=0.95\linewidth]{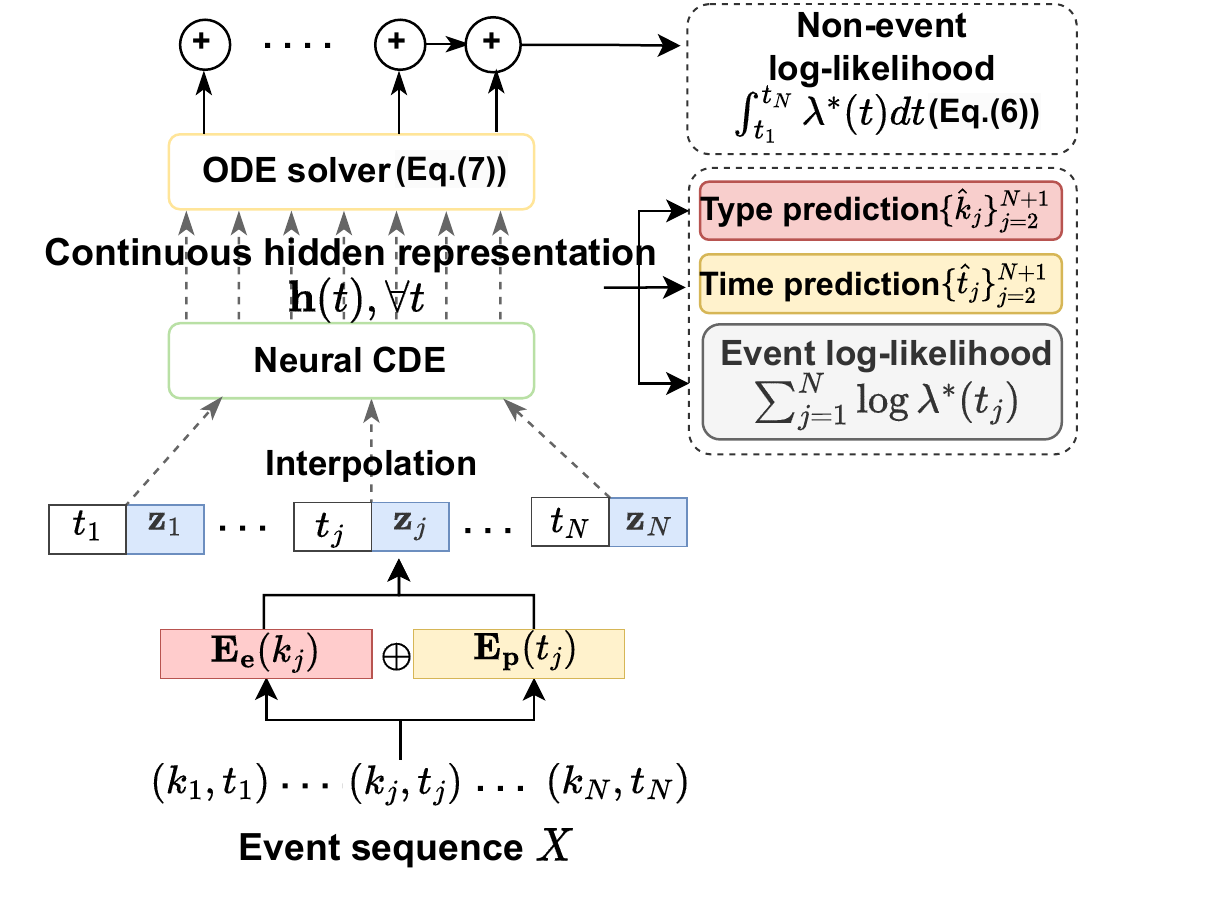}
\end{center}
\caption{Our proposed HP-CDE architecture }
\label{fig:onecol2}
\end{figure}

Figure~\ref{fig:onecol2} shows comprehensive designs of our  proposed model, HP-CDE. The overall workflow is as follows:
\begin{enumerate}
    \item Given the event sequence $X=\{(k_j, t_j)\}_{j=1}^N$, i.e., event type $k_j$ at time $t_j$, the embeddings $\{\mathbf{E_e}(k_j), \mathbf{E_p}(t_j)\}_{j=1}^N$ are made through the encoding processes, where $\mathbf{E_e}(k_j)$ is an embedding of $k_j$ and $\mathbf{E_p}(t_j)$ is a positional embedding of $t_j$.
    \item Then we use  
     $\{\mathbf{E_e}(k_j) \oplus \mathbf{E_p}(t_j)\}_{j=1}^N$ as the discrete hidden representations $\{\mathbf{z}_j\}_{j=1}^N$. In other words, $\mathbf{z}_j = \mathbf{E_e}(k_j) \oplus \mathbf{E_p}(t_j)$, i.e., the element-wise summation of the two embeddings.
    
    \item An interpolation algorithm is used to create the continuous path $Z(t)$ from $\{(\mathbf{z}_j, t_j)\}_{j=1}^N$ --- we augment the time information $t_j$ to each $\mathbf{z}_j$.
    \item Using the continuous path $Z(t)$, a neural CDE layer calculates the final continuous hidden representation $\mathbf{h}(t)$ for all $t$. At the same time, an ODE solver integrates the continuous intensity function $\lambda^*(t)$ which is calculated from $\mathbf{h}(t)$ (cf. Eq.~\eqref{eqn:lambda_prediction}) to calculate the non-event log-likelihood. In addition, there are three prediction layers to predict the event type, time, and log-likelihood (cf. Figure~\ref{fig:onecol3}).
\end{enumerate}

We provide more detailed descriptions for each step in the following subsections with the well-posedness of our model. 

\subsection{Embedding}

We embed both the type and time of each event into separate vectors and then add them. To be more specific, we map each event type to an embedding vector $\mathbf{E_e}(k)$, which is trainable. With trigonometric functions, we embed the time information to a vector $\mathbf{E_p}(t)$, which is called positional encoding in transformer language models (cf. Appendix~\ref{appendix:Embedding}). We use the sum of the two embeddings, $\{\mathbf{E_e}(k_j) \oplus \mathbf{E_p}(t_j)\}_{j=1}^N$ as the discrete hidden representations $\{\mathbf{z}_j\}_{j=1}^N$, i.e., $\mathbf{z}_j = \mathbf{E_e}(k_j) \oplus \mathbf{E_p}(t_j)$.

\subsection{Occurrence Dynamics and Continuous Intensity Function}~\label{sec:occurrence_dynamics}With $\{\mathbf{z}_j\}_{j=1}^N$, we calculate the \emph{continuous} hidden representation $\mathbf{h}(t_j)$ for any arbitrary $j$, where $t_1 \leq t_j$, based on the neural CDE framework as follows:
\begin{linenomath}\begin{align}\label{eqn:NCDE}
\mathbf{h}(t_j) = \mathbf{h}(t_1) + \int_{t_1}^{t_j}f(\mathbf{h}(t);\mathbf{\theta}_f)\frac{dZ(t)}{dt}dt,
\end{align}\end{linenomath}where $Z(t)$ is a continuous path created by an interpolation algorithm from $\{(\mathbf{z}_j, t_j)\}_{j=1}^N$. The well-posedness\footnote{The well-posedness of an initial value problem means that i) its unique solution, given an initial value, exists, and ii) its solutions continuously change as initial values change.} of neural CDEs is proved in \cite[Theorem 1.3]{lyons2004differential} under the Lipschitz continuity requirement (cf. Appendix~\ref{appendix:Well-posedness}). Neural CDE layer is able to generate the continuous hidden representation $\mathbf{h}(t_j)$, where $t_1 \leq t_j$, even when the sequence $\{(\mathbf{z}_j, t_j)\}_{j=1}^N$ is an irregular time-series, i.e., the inter-arrival time varies from one case to another. 

This continuous property enables our model to exactly solve the integral problem of the non-event log-likelihood. That is, the non-event log-likelihood can be re-written as the following ODE form:
\begin{linenomath}\begin{align}\label{eq:exact}
\mathbf{a}(t_N) = \int_{t_1}^{t_N} \lambda^*(t) dt,
\end{align}\end{linenomath}where the conditional intensity function of Eqs.~\eqref{eq:hawkesIntensity} and~\eqref{eq:neuralHawkesIntensity} is, in our case, the sum of the conditional intensity functions of all event types as follows: 
\begin{linenomath}\begin{align}\label{eqn:lambda_prediction}
\lambda^*(t)  &= \sum_{k=1}^K \lambda_k^*(t), \quad \lambda_k^*(t)  = \phi_k(\mathbf{W}^{\text{intst}\top}_k\mathbf{h}(t_j)),
 \end{align}\end{linenomath}
where ${\mathbf{W}^{\text{intst}}_k}$ is a weight matrix of intensity about type $k$, and therefore, ${\mathbf{W}^{\text{intst}\top}_k\mathbf{h}(t_j)}$ is a linear projected representation which has the history of events before time $t_j$. $\phi_k(x) := \beta_k \log (1+ \exp(x/\beta_k))$ is the softplus function with a parameter $\beta_k$ to be learned. The softplus function is used to restrict the intensity function to have only positive values. Therefore, the log-probability of HP-CDE for any event sequence $X$ is redefined from Eq.~\eqref{eq:logProb} as:
\begin{linenomath}\begin{align}\label{eq:newLogProb}
\log p(X) = \sum_{j=1}^{N} \log \lambda^*(t_j) - \mathbf{a}(t_N).
\end{align}\end{linenomath}

As a result, we can naturally define the following augmented ODE, where $\mathbf{h}(t)$ and $\mathbf{a}(t)$ are combined:
\begin{linenomath}\begin{align}\label{eqn:augmented part}
\frac{d}{dt}\begin{bmatrix} \mathbf{h}(t)\\ \mathbf{a}(t) \end{bmatrix}  = 
\begin{bmatrix} 
f(\mathbf{h}(t);\mathbf{\theta}_f)\frac{dZ(t)}{dt} \\ \lambda^*(t) 
\end{bmatrix} 
\end{align}\end{linenomath}
and
\begin{linenomath}\begin{align}\label{eqn:initial value}
\begin{bmatrix} \mathbf{h}(t_1)\\\mathbf{a}(t_1) \end{bmatrix} = 
\begin{bmatrix}
\pi(\mathbf{z}(t_1);\mathbf{\theta}_{\pi}) \\ 
0
\end{bmatrix},
\end{align}\end{linenomath}
where $\pi$ is a fully connected layer. The neural network $f$ is defined as follows:
\begin{linenomath}\begin{align}
   f(\mathbf{h}(t)) = \text{Tanh}(\pi_{M}(\text{ELU}(\cdots (\text{ELU}(\pi_{\text{1}}(\mathbf{h}(t))))))),
\end{align}\end{linenomath}\label{eqn:f_architecture}which consists of fully connected layers with the ELU or the hyperbolic tangent activation. The number of layers $M$ is a hyperparameter.

In Zuo et al.~\cite{10.5555/3524938.3526022}, the generated hidden representations from the self-attention module of their transformer have discrete time stamps, and therefore, its associated intensity function definition is inevitably discrete. For that reason, they rely on a heuristic method, e.g., Monte Carlo method, to calculate the non-event log-likelihood. In our case, however, the physical time is modeled in a continuous manner and therefore, the exact non-event log-likelihood can be calculated as in Eq.~\eqref{eq:exact}.

 \begin{figure}[t] 
\begin{center}

\includegraphics[width=0.7\linewidth]{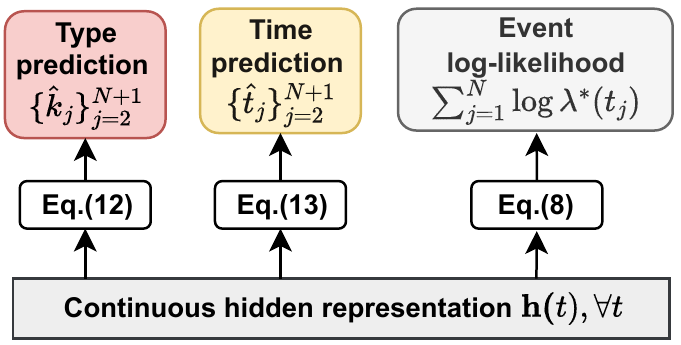}
\end{center}
\label{fig:long}
\caption{Prediction layer of HP-CDE}\label{fig:onecol3}
\end{figure}

\subsection{Prediction Layer}

Our model has three prediction layers as in other Hawkes process models: i) next event type, ii) next event time, and iii) the event log-likelihood (cf. Figure~\ref{fig:onecol3}). We use Eq.~\eqref{eqn:lambda_prediction} to calculate the event log-likelihood.

For the event type and time predictions, we predict $\{\hat t_{j}\}_{j=2}^{N+1}$ and $\{\hat{{k}}_{j}\}_{j=2}^{N+1}$ after reading $X = \{(k_j, t_j)\}_{j=1}^{N}$. For the event type prediction layer, we use the following method:
\begin{linenomath}\begin{align}\begin{split}\label{eqn:type prediction}
\hat{\mathbf{p}}_{j+1} &= \text{Softmax}(\mathbf{W}^{\text{type}}\mathbf{h}(t_j)),\\
\hat{{k}}_{j+1} &= \arg \max_k \hat{\mathbf{p}}_{j+1}(k),
\end{split}\end{align}\end{linenomath}
where $\mathbf{W}^{\text{type}}$ is a trainable parameter and $\hat{\mathbf{p}}_{j+1}(k)$ is the probability of type $k$ at time $t_{j+1}$. For the event time prediction layer, we use the following definition:
\begin{linenomath}\begin{align}\label{eqn:time prediction}
\hat{t}_{j+1} = \mathbf{W}^{\text{time}} \mathbf{h}(t_j),
\end{align}\end{linenomath} where $\mathbf{W}^{\text{time}}$ is a trainable parameter.

\subsection{Training Algorithm}

Our loss definition consists of three parts. The first part is the following MLE loss, i.e. maximizing the log-likelihood (cf. Eq.~\eqref{eq:newLogProb}):
\begin{linenomath}\begin{align}
\max \sum_{i=1}^S \log p(X_i),
\end{align}\end{linenomath}
where $S$ is the number of training samples. While training, the log-intensity of each observed event increases and the non-event log-likelihood decreases in the whole interval $[t_1, t_N]$.

The second loss is the event type loss function which is basically a cross-entropy term as follows:
\begin{linenomath}\begin{align}\label{eqn:type loss}
\mathcal{L}_{\text{type}}(X) = \sum_{j=2}^{N+1}  -\mathbf{k}_j^\top \log(\hat{\mathbf{p}}_{j}),
\end{align}where $\mathbf{k}_j$ is a one-hot vector for the
event type $k_j$. In the case of the event time loss, we use the inter-arrival time $\tau_i=t_i-t_{i-1}$ to compute the loss as follows:
\begin{align}\label{eqn:time loss}
\mathcal{L}_{\text{time}}(X) = \sum_{j=2}^{N+1} (\tau_{j} - \hat{\tau}_{j})^2.
\end{align}\end{linenomath}

Therefore, the overall objective function of HP-CDE can be written as follows:
\begin{linenomath}\begin{align}\label{eqn:loss}
\min  \sum_{i=1}^S- \alpha_1\log p({X_i}) +  \mathcal{L}_{\text{type}}(X_i) + \alpha_2 \mathcal{L}_{\text{time}}(X_i),
\end{align}\end{linenomath}
where $\alpha_1$ and $\alpha_2$ are hyperparameters.

In Alg.~\eqref{alg:train}, we show the training algorithm. We first initialize all the parameters. From our training data, we randomly build a mini-batch $\{X_i\}_{i=1}^{S}$ in Line~\ref{a:mini} --- the optimal mini-batch size varies from one dataset to another. After feeding the constructed mini-batch into our model, we calculate the discrete and continuous hidden representations in Lines~\ref{a:disc} and~\ref{a:cont}. With the loss in Eq.~\eqref{eqn:loss}, we train our model. We repeat the steps $max\_iter$ times.

\begin{algorithm}[t]
\caption{How to train HP-CDE}
\label{alg:train}
    \textbf{Input}: Training data $\mathcal{D}_{train}$, Iteration numbers $max\_iter$ \\
    \begin{algorithmic}[1]
        \STATE Initialize all the parameters of the embedding and the neural CDE layer 

        \STATE $iter \gets 0$
        \WHILE {$iter < max\_iter$}
        \STATE Sample a mini-batch $\{X_i\}_{i=1}^{S} \in \mathcal{D}_{train}$\label{a:mini}
    
        \STATE Calculate the embedding vectors, i.e, $\mathbf{E_e}(k_j)$, and $\mathbf{E_p}(t_j)$
    
        \STATE Calculate the discrete hidden representation $\mathbf{z}_j, \forall j$\label{a:disc}

        \STATE Calculate the continuous hidden representation $\mathbf{h}(t)$ using neural CDE and compute the non-event log-likelihood using ODE solver with Eq.~\eqref{eq:exact} over time\label{a:cont}

        \STATE Update the parameters with Eq.~\eqref{eqn:loss}
        
        \IF {the loss does not decrease for $\delta$ iterations} 
        \STATE exit
        \ENDIF
    \ENDWHILE

\STATE \textbf{return} the trained parameters
\end{algorithmic}
\end{algorithm}

\begin{table*}[t]
\setlength{\tabcolsep}{7pt}
\centering
\small

\begin{tabular}{c|c|rrr|rr}
\Xhline{3\arrayrulewidth}
\multirow{2}{*}{Dataset} & \multirow{2}{*}{Model} & \multirow{2}{*}{LL $\uparrow$} & \multirow{2}{*}{ACC $\uparrow$} & \multirow{2}{*}{RMSE $\downarrow$}  & Memory & Training \tabularnewline
& & &  & &   usage(MB) & time(m)  
\tabularnewline
\Xhline{3\arrayrulewidth}
\multirow{5}*{MIMIC} 
& RMTPP        & -1.222\footnotesize{$\pm$0.080} & 0.823\footnotesize{$\pm$0.014} & 1.035\footnotesize{$\pm$0.023}  & 3 & 0.004\tabularnewline
& NHP          & -0.647\footnotesize{$\pm$0.051} & 0.534\footnotesize{$\pm$0.015} & 0.976\footnotesize{$\pm$0.020} & 13 & 0.045 \tabularnewline
& SAHP         & -0.859\footnotesize{$\pm$0.328} & 0.555\footnotesize{$\pm$0.171} & 1.138\footnotesize{$\pm$0.059}   & 34 & 0.037 \tabularnewline

& THP & -0.233\footnotesize{$\pm$0.012} & 0.741\footnotesize{$\pm$0.021} & 0.856\footnotesize{$\pm$0.040}   & 9 & 0.012\tabularnewline\cline{2-7}
& HP-CDE     & \textbf{2.573\footnotesize{$\pm$0.201}} & \textbf{0.847\footnotesize{$\pm$0.007}} & \textbf{0.726\footnotesize{$\pm$0.042}}   & 58& 0.058\tabularnewline

\Xhline{2\arrayrulewidth}
\multirow{5}*{MemeTracker}
& RMTPP        & NaN &  0.006\footnotesize{$\pm$0.000} & NaN  & 1,708 &  0.425\tabularnewline
& NHP          & -9.395\footnotesize{$\pm$2.814} & 0.044\footnotesize{$\pm$0.003} & 441.293\footnotesize{$\pm$0.233}  & 5,096 & 12.263\tabularnewline
& SAHP         & 2.160\footnotesize{$\pm$0.324} & 0.009\footnotesize{$\pm$0.000} & 521.672\footnotesize{$\pm$4.071}  & 32,894 & 6.642\tabularnewline

& THP  & -5.717\footnotesize{$\pm$0.649} & 0.015\footnotesize{$\pm$0.000} & 446.477\footnotesize{$\pm$2.665}  & 891 & 2.610\tabularnewline\cline{2-7}
& HP-CDE     & \textbf{3.846\footnotesize{$\pm$0.626}} & \textbf{0.151\footnotesize{$\pm$0.005}} & \textbf{441.223\footnotesize{$\pm$3.480}} & 3,669& 3.817\tabularnewline

\Xhline{2\arrayrulewidth}
\multirow{5}*{Retweet}
& RMTPP        & NaN &  0.490\footnotesize{$\pm$0.000} & NaN   & 210 &  0.044 \tabularnewline
& NHP          & -9.082\footnotesize{$\pm$0.125} & 0.547\footnotesize{$\pm$0.010} & 16,630.956\footnotesize{$\pm$0.217}  & 750 & 17.820\tabularnewline
& SAHP         & 1.904\footnotesize{$\pm$0.566} & 0.505\footnotesize{$\pm$0.067} & 16,648.339\footnotesize{$\pm$1.436}   & 13,276 & 0.197\tabularnewline

& THP   & -7.347\footnotesize{$\pm$0.268} & 0.499\footnotesize{$\pm$0.013} & \textbf{15,050.470\footnotesize{$\pm$26.712} }  & 1,582 & 0.142\tabularnewline\cline{2-7}
& HP-CDE     & \textbf{6.844\footnotesize{$\pm$0.539}} & \textbf{0.552\footnotesize{$\pm$0.009}} & 15,849.218\footnotesize{$\pm$269.068}  & 197 & 6.236\tabularnewline

\Xhline{2\arrayrulewidth}

\multirow{5}*{StackOverFlow}
& RMTPP        & -1.894\footnotesize{$\pm$0.002} & 0.429\footnotesize{$\pm$0.000} & 1.321\footnotesize{$\pm$0.002}   & 27 & 0.040\tabularnewline
& NHP          & -7.726\footnotesize{$\pm$0.581} & 0.434\footnotesize{$\pm$0.015} & 1.027\footnotesize{$\pm$0.027}   & 449 & 3.556\tabularnewline
& SAHP         & -0.431\footnotesize{$\pm$0.225} & 0.244\footnotesize{$\pm$0.002} & 4.525\footnotesize{$\pm$1.098}  & 11,080 & 0.147 \tabularnewline

& THP  & -0.554\footnotesize{$\pm$0.001} & 0.449\footnotesize{$\pm$0.001} & \textbf{0.973\footnotesize{$\pm$0.001}}  & 4,585 & 0.169 \tabularnewline\cline{2-7}
& HP-CDE     & \textbf{7.348\footnotesize{$\pm$0.466}} & \textbf{0.452\footnotesize{$\pm$0.001}} & 0.996\footnotesize{$\pm$0.017} & 44 & 6.878\tabularnewline

\Xhline{3\arrayrulewidth}
\end{tabular}
\caption{Experimental results.$\uparrow$ (resp. $\downarrow$) denotes that the higher (resp. lower) the better, and we use boldface to denote the best score.}~\label{tbl:mainResults}
\end{table*}

\section{Experiments} 
\subsection{Experimental Environments}

\subsubsection{Experimental Settings}
In this section, we compare the model performance of HP-CDE with 4 state-of-the-art baselines on 4 datasets. Each dataset is split into the training set and the testing set. The training set is used to tune the hyperparameters and the testing set is used to measure the model performance. We evaluate the models with three metrics: i) log-likelihood (LL) of $X = \{(k_j, t_j)\}_{j=1}^N$, ii) accuracy (ACC) on the event type prediction, and iii) root mean square error (RMSE) on the event time prediction. 
We train each model 100 epochs and report the mean and standard deviation of the evaluation metrics of five trials with different random seeds. We compare our model with various baselines (cf. Section~\ref{sec:hp_history}): Recurrent Marked Temporal Point Process (RMTPP)\footnote{https://github.com/dunan/NeuralPointProcess}, Neural Hawkes Process (NHP)\footnote{https://github.com/hongyuanmei/neural-hawkes-particle-smoothing},  Self-Attentive Hawkes Process (SAHP)\footnote{https://github.com/QiangAIResearcher/sahp\_repo}, and Transformer Hawkes Process (THP)\footnote{https://github.com/SimiaoZuo/Transformer-Hawkes-Process}. 
More details including hyperparameter configurations are in Appendix~\ref{appendix:experimental_details}.

\subsubsection{Datasets}
To show the efficacy and applicability of our model, we evaluate using various real-world data.
MemeTracker~\cite{snapnets}, Retweet~\cite{10.1145/2783258.2783401}, and StackOverFlow~\cite{snapnets}, are collected from Stackoverflow, web articles, and Twitter, respectively. We also use a medical dataset, called MIMIC~\cite{mimic3}. We deliberately choose the datasets with various average sequence lengths and event type numbers $K$ to show the general efficacy of our model. The average sequence length ranges from 3 to 109, and the number of event types $K$ ranges from 3 to 5000 (cf. Table~\ref{tbl:dataset}). That is, we cover not only from simple to complicated ones, but also from short-term to long-term sequences. {Details of datasets are in Appendix~\ref{appendix:dataset}}

\begin{table}[t]
\setlength{\tabcolsep}{6pt}
\small
\centering

\begin{tabular}{c|r|rrr|r}
\Xhline{3\arrayrulewidth}
\multirow{2}{*}{Dataset}  & \multirow{2}{*}{$K$}     & \multicolumn{3}{c|}{Sequence length} &  \multirow{2}{*}{\# Events}  \\
\cline{3-5} 
 &      & Min  & Average  & Max &                     \\\Xhline{3\arrayrulewidth}

 MIMIC         & 75   & 2    & 4       & 26               & 1,930  \\
 MemeTracker   & 5000 & 1    & 3       & 31               & 123,639  \\
 Retweet       & 3    & 50   & 109     & 264              & 2,173,533 \\
 StackOverFlow & 22   & 41   & 72      & 720              & 345,116 \\
\Xhline{3\arrayrulewidth}
\end{tabular}
\caption{Characteristics of datasets used in experiments}\label{tbl:dataset}
\end{table}


\subsection{Experimental Results}We show the experimental results of each model on MIMIC, MemeTracker, Retweet, and StackOverFlow in Table~\ref{tbl:mainResults}. We analyze the results in three aspects: i) the event prediction, ii) the log-likelihood, and iii) the model complexity. Ablation and sensitivity analyses are in Appendix~\ref{appendix:sensitivity} and~\ref{appendix:ablation}.

\begin{table}[t]
\setlength{\tabcolsep}{6pt}
\centering
\small

\begin{tabular}{c|r|r}
\Xhline{3\arrayrulewidth}
\multirow{2}{*}{Model} &  \multicolumn{2}{c}{Dataset}  \\ \cline{2-3}
& MIMIC & MemeTracker \\ \Xhline{3\arrayrulewidth}
RMTPP & 0.385\footnotesize{$\pm$0.037} & 0.000\footnotesize{$\pm$0.000} \\
NHP & 0.126\footnotesize{$\pm$0.018}& 0.011\footnotesize{$\pm$0.002}\\
SAHP & 0.108\footnotesize{$\pm$0.112}& 0.000\footnotesize{$\pm$0.000}\\
THP & 0.162\footnotesize{$\pm$0.016}& 0.000\footnotesize{$\pm$0.000}\\ \cline{1-3}

HP-CDE & \textbf{0.452\footnotesize{$\pm$0.035}}& \textbf{0.069\footnotesize{$\pm$0.004}}\\
\Xhline{3\arrayrulewidth}
\end{tabular}
\caption{F1 score ($\uparrow$) for imbalanced datasets}~\label{tbl:f1score}
\end{table}

\subsubsection{Event Prediction}
HP-CDE outperforms other baselines with regards to both the event type and the event time prediction in most cases as reported in Table~\ref{tbl:mainResults}. To be specific, in terms of accuracy,  HP-CDE shows the best performance in every dataset. These results imply that processing data in a continuous manner is important when it is in a continuous time domain. Even though HP-CDE only shows the lowest RMSE on datasets with short sequnce length, MIMIC and MemeTracker, we provide the solution to lower RMSE of HP-CDE when using datasets with long sequence length in Section~\ref{sec:ablation}.

\begin{figure}[t]
    \centering
    
    \subfigure[MIMIC]{\includegraphics[width=0.48\columnwidth]{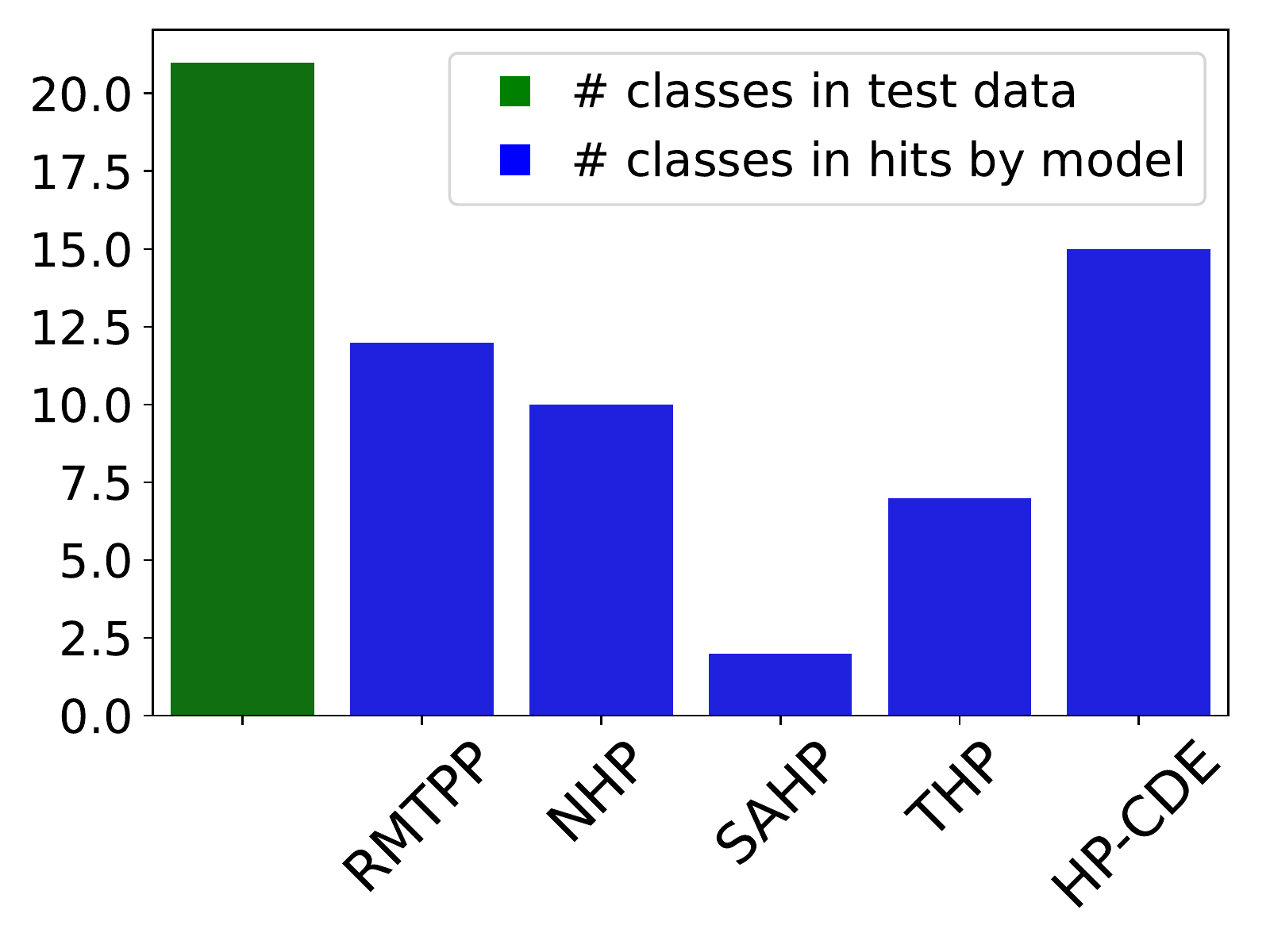}}
    \subfigure[MemeTracker]{\includegraphics[width=0.48\columnwidth]{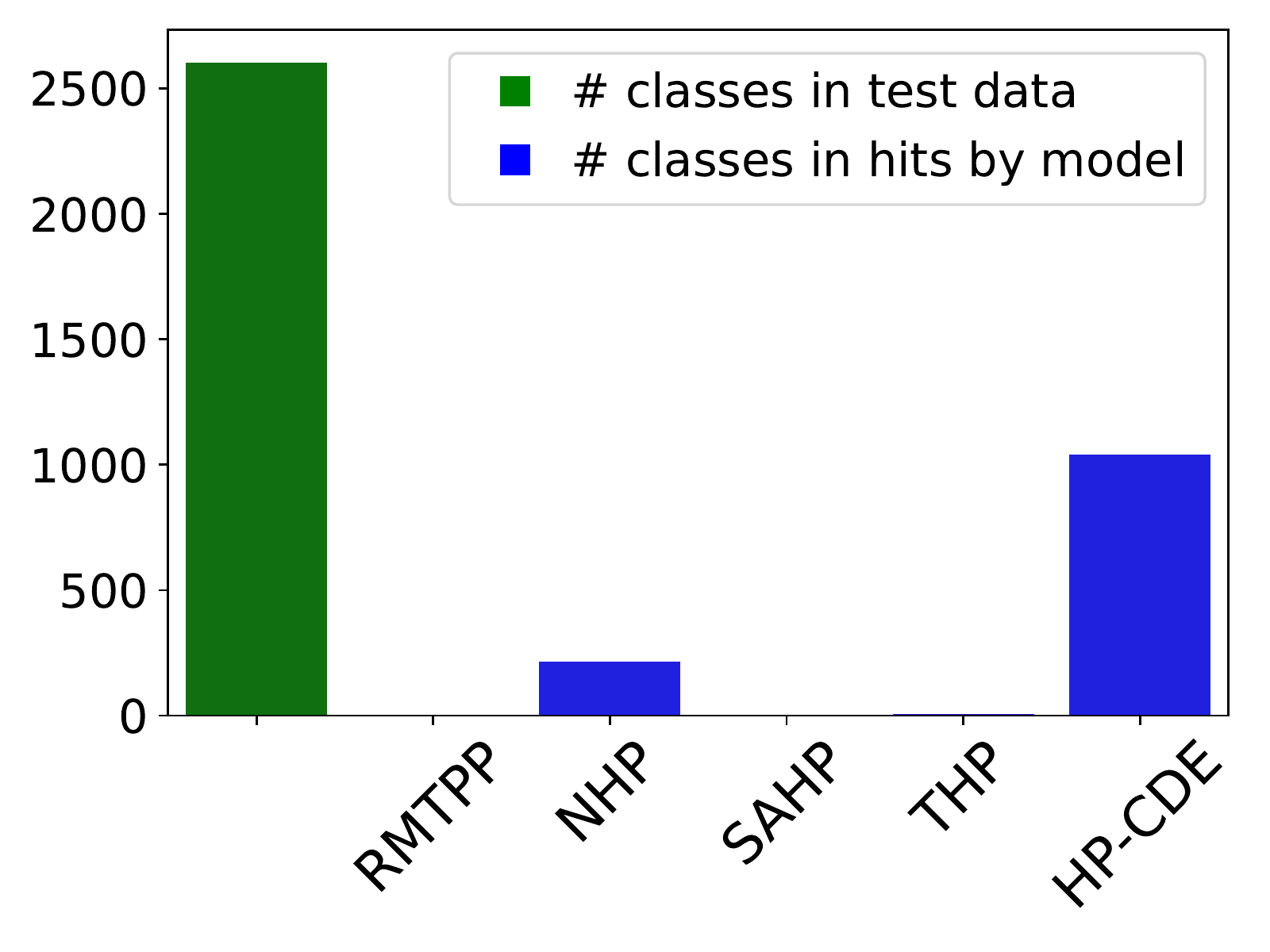}}
    \caption{The number of classes in test data vs. the number of classes in correct event type predictions, i.e., hits. HP-CDE provides not only accurate but also diverse predictions.}
    
    \label{fig:class}
\end{figure}

For the imbalanced datasets of MIMIC and MemeTracker, where only 20\% of types occupy 90\% and 70\% of events each, we do the following additional analyses. Notably, HP-CDE attains an accuracy of 0.151 in MemeTracker, which is up to 243\% higher than those of baselines, and an RMSE of 0.726 in MIMIC, about 15\% lower. Furthermore, we use the macro F1 score to measure the quality of type predictions. As shown in Table~\ref{tbl:f1score}, our model shows the best F1 score in both of the imbalanced datasets. Especially for MemeTracker, models with attention modules have relatively low F1 scores, indicating that when there exist too many classes and if they are imbalanced, attentions are overfitted to several frequently occurring classes. This phenomenon is also observed in Figure~\ref{fig:class}. In Figure~\ref{fig:class}, HP-CDE shows the most diverse predictions in terms of the number of predicted classes. Particularly, in Figure~\ref{fig:class} (b), HP-CDE successfully predicts for 1,164 classes among 2,604 classes, which is almost 50\% of the classes in test data, whereas NHP, SAHP, and THP predict only for 217, 4, and 7 classes, respectively.

Regardless of the characteristics of datasets, e.g., the number of types, the degree of imbalance, and so on, our model shows outstanding prediction results, which prove the importance of continuous processing and computing the exact log-likelihood leading to more accurate learning of dynamics.

\subsubsection{Log-likelihood Calculation}As shown in Table~\ref{tbl:mainResults}, our models always show the best log-likelihood, outperforming others by large margins, on every dataset. One remarkable point is that our log-likelihood is always positive, while baselines show negative values in many cases. That is, in HP-CDE, the event log-likelihood exceeds the non-event log-likelihood at all times.

\begin{figure}[t]
    \centering
    \subfigure[Retweet]{\includegraphics[width=0.48\columnwidth]{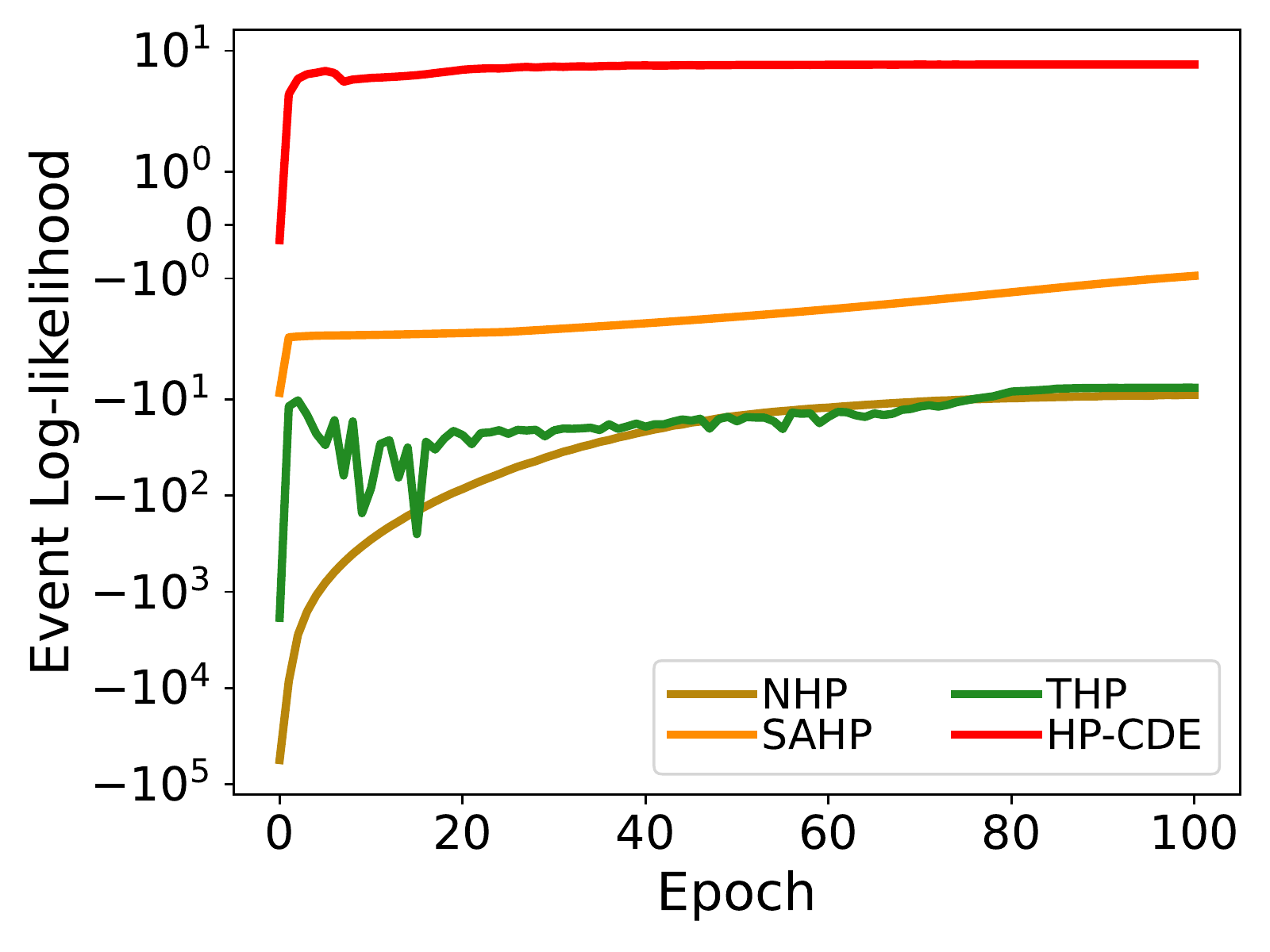}}
    \subfigure[MemeTracker]{\includegraphics[width=0.48\columnwidth]{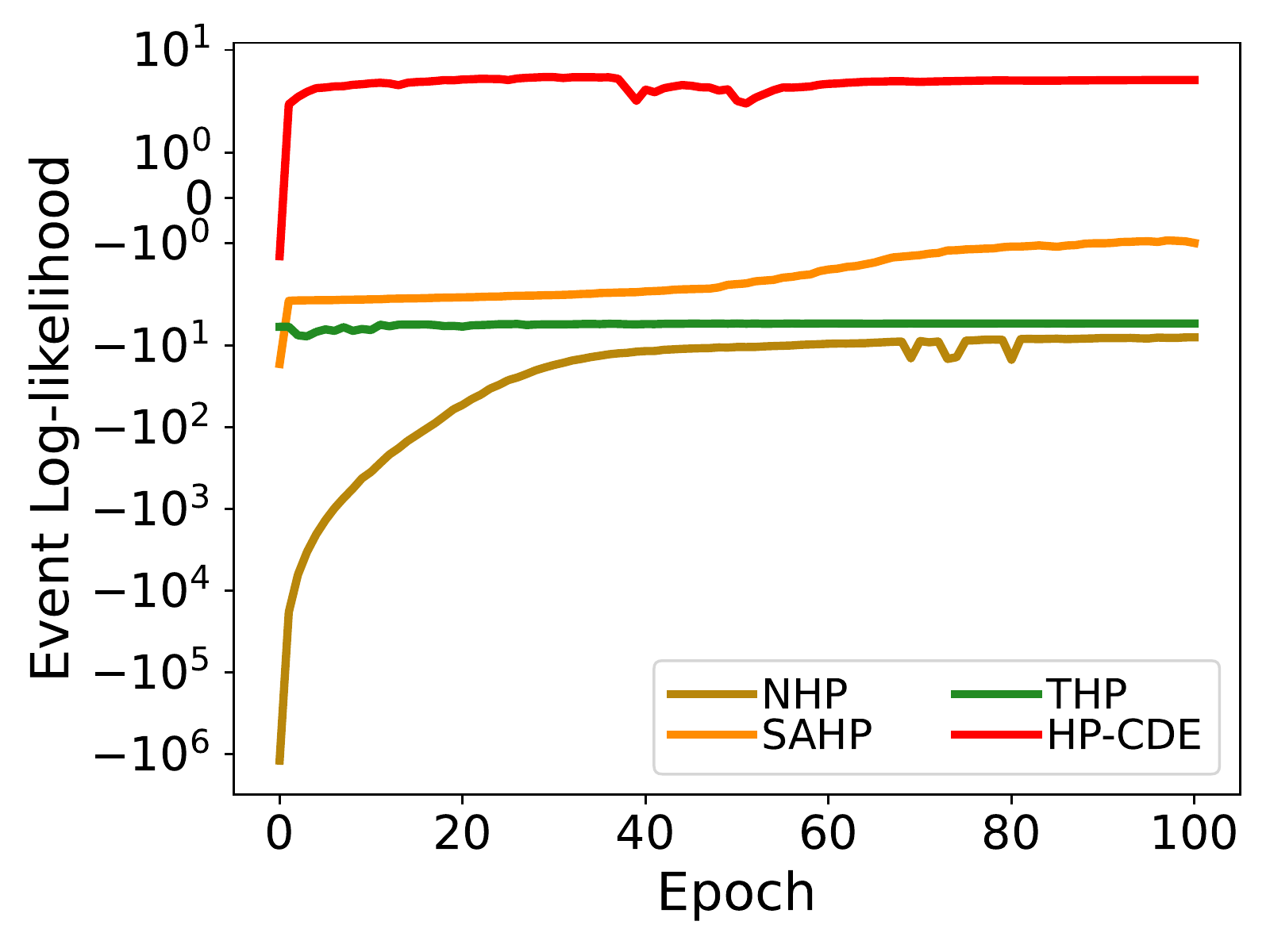}}
    \caption{Training curves on Retweet and MemeTracker. HP-CDE shows the highest log-likelihood with the fastest convergence speed.}
    \label{fig:ll_tracking}
\end{figure}

Figure~\ref{fig:ll_tracking} shows the training curves of models fitted on {Retweet and MemeTracker in a log-scale. First of all, HP-CDE show the best log-likelihood at every training epoch. Overall, except THP, the log-likelihood of MemeTracker tends to be more unstable than that of Retweet, since MemeTracker has about 1,700 times more event types than Retweet.} 

\subsubsection{Memory Usage}Table~\ref{tbl:mainResults} also recaps the model complexity. {Exactly calculating the non-event log-likelihood using ODE solvers incurs additional memory usage, so that the model uses bigger memory than those of other sampling methods such as Monte Carlo sampling.} Especially when the number of event types $K$ is large, i.e., MIMIC and MemeTracker, the complexity of HP-CDEs increases as we exactly compute the non-event log-likelihood for every event type. However, when $K$ is relatively small, owing to the adjoint sensitivity method~\cite{NIPS2018_7892,kidger2020neural}, HP-CDE's memory footprint notably decreases. For example, when using Retweet with $K=3$, the space complexity of HP-CDE is almost 1\% of that of THP.

\begin{figure}[t]
    \centering
    
    \subfigure[Retweet]{\includegraphics[width=0.48\columnwidth]{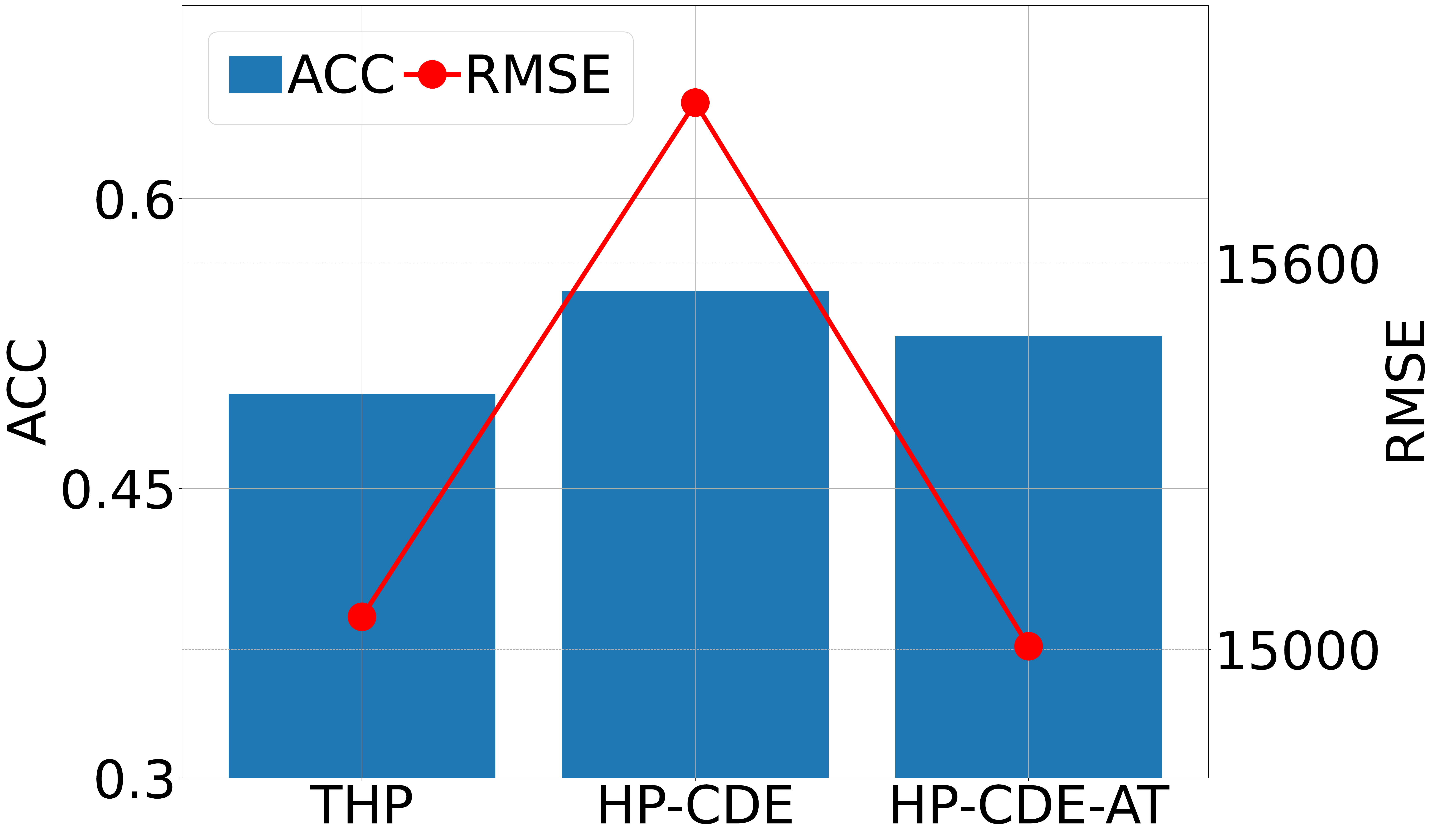}}
    \subfigure[StackOverFlow]{\includegraphics[width=0.48\columnwidth]{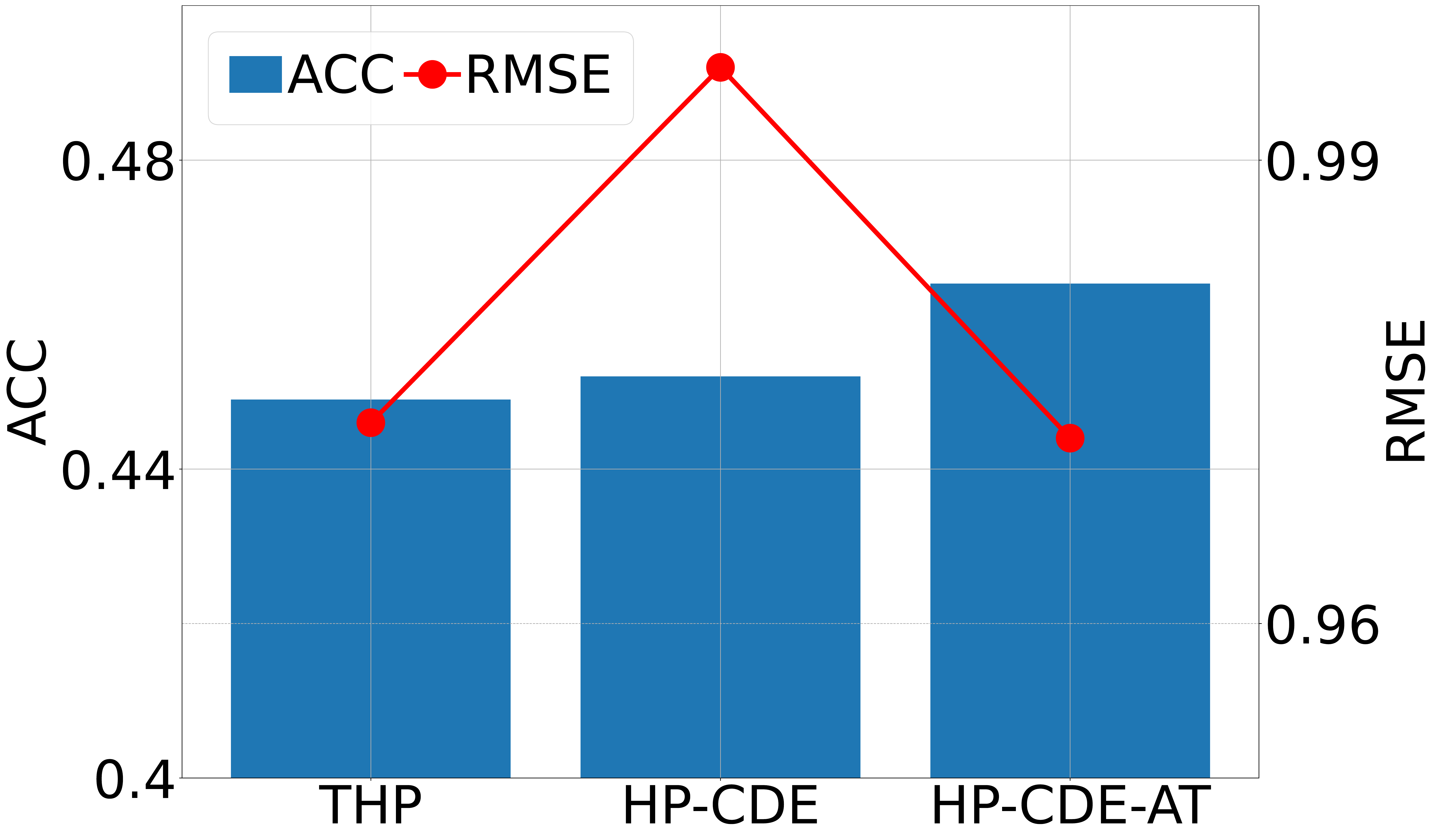}}
    \caption{Additional study on long-sequence datasets, comparing accuracy and RMSE of HP-CDE-AT to HP-CDE and THP.}
    
    \label{fig:ablation_long}
\end{figure}

\subsection{Additional Study on the Long Sequence Length}\label{sec:ablation}
While HP-CDE shows a good performance on the datasets with relatively short sequence lengths, i.e., MIMIC and MemeTracker, its RMSE results on others with longer sequence lengths, i.e., Retweet and StackOverFlow, are slightly larger than those of THP's.  Therefore, to effectively deal with long sequence datasets, we put the self-attention part of transformer~\cite{NIPS2017_3f5ee243} right before the neural CDE layer and name the model HP-CDE-AT. Experimental results of HP-CDE-AT in comparison with HP-CDE and THP, which shows the highest score among baselines, are summarized in Figure~\ref{fig:ablation_long}. According to Figure~\ref{fig:ablation_long} (a), HP-CDE-AT achieves the smallest RMSE, improving the performance of the origianl HP-CDE model. Remarkably, in Figure~\ref{fig:ablation_long} (b), HP-CDE-AT even shows the best performance on StackOverFlow in both metrics, accuracy and RMSE. In conclusion, since HP-CDE-AT attains overall best results on longer datasets, HP-CDE-AT is one good option for long sequence datasets (cf. Appendix~\ref{appendix:HP-CDE-AT}).


\section{Conclusions}Temporal point processes are frequently used in real-world applications to model occurrence dynamics in various fields. In particular, deep learning-based Hawkes process models have been extensively studied. However, we identified the two possible enhancements from the literature and presented HP-CDE to overcome the limitations. First, we use neural CDEs to model occurrence dynamics since one of their main application areas is to model uncertainties in human behaviors. Second, we exactly calculate the non-event log-likelihood which is one important part of the training objective. Existing work uses heuristic methods for it, which makes the training process unstable sometimes. In our experiments, consequently, our presented method significantly outperforms them and shows the most diverse predictions, i.e., the least overfitting.

\section*{Acknowledgements}

Noseong Park is the corresponding author. This work was supported by the Institute of Information \& Communications Technology Planning \& Evaluation (IITP) grant funded by the Korean government (MSIT) (No. 2020-0-01361, Artificial Intelligence Graduate School Program at Yonsei University, 10\%), and (2022-0-01032, Development of Collective Collaboration Intelligence Framework for Internet of Autonomous Things, 45\%) and (No.2022-0-00113, Developing a Sustainable Collaborative Multi-modal Lifelong Learning Framework, 45\%).

\section*{Ethical Statement} 
MIMIC contains much personal health information. However, it was released after removing observations, such as diagnostic reports and physician notes, using a rigorously evaluated deidentification system to protect the privacy of the patients who have contributed their information. Therefore, our work does not have any related ethical concerns.

\bibliographystyle{named}
\bibliography{ijcai23}

\clearpage


\appendix

\section{Positional Encoding}~\label{appendix:Embedding} 
With the sinusoidal and cosine functions, we embed the temporal information to a vector $\mathbf{E_p}(t)$, which we call positional encoding: 
\begin{linenomath}
\begin{align}\label{eq:temporal encoding}
[\mathbf{E_p}(t_j)]_u\begin{cases} \cos(t_j/10000^{\frac{u-1}{\dim(\mathbf{z}_j)}}) , \quad \text{if }u \text{ is odd,} \\ \sin(t_j/10000^{\frac{u}{\dim(\mathbf{z}_j)}}) , \quad \text{if }u \text{ is even.}\end{cases}
\end{align}
\end{linenomath}
$[\mathbf{E_p}(t_j)]_u$ denotes the $u$-th element of the embedding vector obtained from the temporal positional encoding of $t_j$ in the sequence $X$. The sinusoidal and cosine functions let our model be able to process the sequence length that are longer than those encountered during training~\cite{NIPS2017_3f5ee243,10.5555/3524938.3526022,pmlr-v119-zhang20q}. Due to the characteristic of the functions, the encoding scheme reflects the relative positional information of the events into the embeddings. Therefore, applying positional encoding can especially bring more advantages when each sequence has a different sequence length, which is highly likely in real-world point process applications.

\section{Well-posedness}~\label{appendix:Well-posedness}
The initial value problem of our model in Eq.~\eqref{eqn:NCDE} is well-posed since we use Lipschitz-continuous operations to construct the neural network $f$. For example, batch normalization, dropout and other pooling techniques, which are usual neural network layers, have explicit Lipschitz constant values. On top of that, most of the activations, such as ELU, ReLU, Tanh, ArcTan, Sigmoid, and Softsign, have a Lipschitz constant of 1. Consequently, in every case, this property ensures that a unique solution exists and that our training process to be stable.

\section{Experimental Details}~\label{appendix:experimental_details}
\subsection{Environments}~\label{appendix:environments}Our software and hardware environments are as follows: \textsc{Ubuntu} 18.04 LTS, \textsc{Python} 3.9.7, \textsc{Numpy} 1.21.2, \textsc{Scipy} 1.7.3, \textsc{Matplotlib} 3.5.1,  \textsc{CUDA} 11.6, and \textsc{NVIDIA} Driver 510.85.02, i9 CPU, and \textsc{NVIDIA RTX A6000}.

\subsection{Hyperparameters}~\label{appendix:hyperparameter}A batch size $S$ of 16 is used for MIMIC and StackOverFlow, 128 for Retweet, and 512 for MemeTracker. For the baselines, we follow the best hyperparameter sets reported in the papers. In the case of HP-CDE, we adopt the Adam optimizer with early stopping and search the hyperparameters as follows (cf. Table ~\ref{tbl:appendix_parameter}):

     We set a weight decay to $1.0\times 10^{-5}$ and use a learning rate of $\{1.0\times10^{-3}, 5.0\times10^{-3}\}$, an embedding size $\dim(\mathbf{z}_j)$ of $\{50,60,70,80\}$, $\alpha_1$ of $\{1.0\times10^{-4},1.0\times10^{-1},1.0\times10^{-0}\}$, and $\alpha_2$ of $\{1.0\times10^{-4}, 1.0\times10^{-3}, 1.0\times10^{-2}\}$. With respect to the hyperparameters of the neural CDE function $f$, we search $\{4,5,6\}$ for $M$, $\{16,32,64,96,128\}$ for the size of the hidden vector $\mathbf{h}(t)$, i.e., $\dim(\mathbf{h})$, and $\{15, 45, 90\}$ for the size of $\pi_m$, where $m=\{1,\cdots,M\}$. For the early stopping, we use $\delta = 5$.


\subsection{Datasets}~\label{appendix:dataset}
\vspace{-0.6cm}
\subsubsection{Descriptions}
We present detailed descriptions of four datasets that are used in our experiments as follows:

\begin{enumerate}
\item MIMIC ~\cite{mimic3}: The Multiparameter Intelligent Monitoring in Intensive Care (MIMIC) dataset contains electrical medical records of patients' diagnoses from 2001 to 2008. There are 75 types of diagnoses and the time stamp of visits is served as event time to learn the patients' occurrence dynamics.

\item MemeTracker~\cite{snapnets}: The MemeTracker dataset consists of many event sequences with many event types, i.e., $K$=5000. The dataset is collected from over 1.5 million documents such as web articles. Each event type stands for user id who used a certain meme, and each event time is a corresponding timestamp.

\item Retweet~\cite{snapnets}: The Retweet dataset consists of retweet sequences from Twitter with 3-type retweeters depending on the volume of followers, i.e., ``small'', ``medium'', and ``large''.

\item StackOverFlow~\cite{snapnets}: This dataset is obtained from StackOverFlow, which is a question-and-answer website with an awarding system. Thus, the StackOverFlow dataset contains 22-type awards, e.g., Good Answer, Famous Question, etc., as an event type and the awarded time as an event time.

\end{enumerate}

\subsubsection{Degree of imbalance}

\begin{table}[h]
\setlength{\tabcolsep}{6pt}
\small
\centering

\begin{tabular}{c|r|ccc}
\Xhline{3\arrayrulewidth}
\multirow{2}{*}{Dataset}  & \multirow{2}{*}{$K$}     & \multicolumn{3}{c}{Ratio (\%)} \\
\cline{3-5} 
 &      & Top 1  & Top 20\%  & Top 50\%                 \\\Xhline{3\arrayrulewidth}

 MIMIC         & 75   & 32.59    & 89.33       & 97.10                \\
 MemeTracker   & 5000 & 0.96    & 68.98       & 91.08                \\
 Retweet       & 3    & -   & 49.42     & 95.39               \\
 StackOverFlow & 22   & 43.53   & 78.89      & 96.56               \\
\Xhline{3\arrayrulewidth}
\end{tabular}
\caption{Event type imbalance}\label{tbl:imbalance}
\end{table}

We summarize the degree of imbalance on each dataset in Table~\ref{tbl:imbalance} with three indices. The first index, i.e. Top 1, is the frequency (in percentage) of the most frequent event. Top 20\%/50\% indicates the percentages of the top 20\%/50\% event types. Since Retweet has only 3 types, we make an exception for Retweet, so that Top 20\% and Top 50\% indicate the top 1 and top 2 frequently occurred types, respectively. As shown in Table~\ref{tbl:imbalance}, all the datasets are severely imbalanced, as only half of the types occupy almost the whole of the events, which is more than 90\%. Particularly in MIMIC, we can find that only 20\% of types cause almost 90\% of events, which is significantly imbalanced but reasonable when considering that it is a medical diagnosis dataset. Likewise, most of the datasets used in Hawkes process are imbalanced and therefore, the model's capability to prevent overfitting is one important issue.

\begin{table*}[h]
\setlength{\tabcolsep}{7pt}
\centering
\small

\begin{tabular}{c|c|ccccccc}
\Xhline{3\arrayrulewidth}
Model & Dataset & Learning rate & $\dim(\mathbf{z}_j)$ & $\alpha_1$  & $\alpha_2$ & $M$ & $\dim(\mathbf{h})$ & Hidden size of $\pi_m$ \tabularnewline

\Xhline{3\arrayrulewidth}
\multirow{5}*{HP-CDE}
& MIMIC           & $1.0\times10^{-3}$ & 70 & $1.0\times10^{-1}$ & $1.0\times10^{-2}$ & 6 & 128 & 90 \tabularnewline
& MemeTracker     & $1.0\times10^{-3}$ & 70 & $1.0\times10^{-4}$ & $1.0\times10^{-4}$ & 5 & 64  & 15 \tabularnewline
& Retweet         & $5.0\times10^{-3}$ & 80 & $1.0\times10^{-4}$ & $1.0\times10^{-4}$ & 4 & 16  & 15 \tabularnewline
& StackOverFlow   & $5.0\times10^{-3}$ & 50 & $1.0\times10^{-0}$ & $1.0\times10^{-2}$ & 4 & 32  & 15 \tabularnewline
\Xhline{3\arrayrulewidth}
\end{tabular}
\caption{The best hyperparameter configuration of HP-CDE}~\label{tbl:appendix_parameter}
\end{table*}

\begin{table*}[t]
\setlength{\tabcolsep}{7pt}
\centering
\small

\begin{tabular}{c|c|ccc}
\Xhline{3\arrayrulewidth}

\multirow{2}{*}{Dataset} & \multirow{2}{*}{Method} & \multirow{2}{*}{LL $\uparrow$} & \multirow{2}{*}{ACC $\uparrow$} & \multirow{2}{*}{RMSE $\downarrow$}  \tabularnewline
                 &  &  &  &  \tabularnewline
\Xhline{3\arrayrulewidth}

\multirow{2}*{MIMIC} 
& Monte Carlo  &  0.058\footnotesize{$\pm$0.006}
               &  0.839\footnotesize{$\pm$0.008}
               &  0.729\footnotesize{$\pm$0.006}
               \\
& ODE solver  & \textbf{2.573\footnotesize{$\pm$0.201}} 
              & \textbf{0.847\footnotesize{$\pm$0.007}} 
              & \textbf{0.726\footnotesize{$\pm$0.042}} 
               \\
              
\Xhline{3\arrayrulewidth}
\multirow{2}*{StackOverFlow} 
& Monte Carlo  &  -0.882\footnotesize{$\pm$0.181}
               &  0.452\footnotesize{$\pm$0.001}
               &  1.012\footnotesize{$\pm$0.037}
               \\ 
& ODE solver  & \textbf{7.348\footnotesize{$\pm$0.466}} 
              & \textbf{0.452\footnotesize{$\pm$0.001}} 
              & \textbf{0.996\footnotesize{$\pm$0.017}} 
              \\
              
\Xhline{3\arrayrulewidth}

\end{tabular}
\caption{Comparison of the Monte Carlo method vs. our proposed exact method for calculating the event log-likelihood}~\label{tbl:ablation}

\end{table*}

\section{Sensitivity Study}\label{appendix:sensitivity}

\begin{figure}[t]
    \centering
    
    \subfigure[Log-likelihood]{\includegraphics[width=0.48\columnwidth]{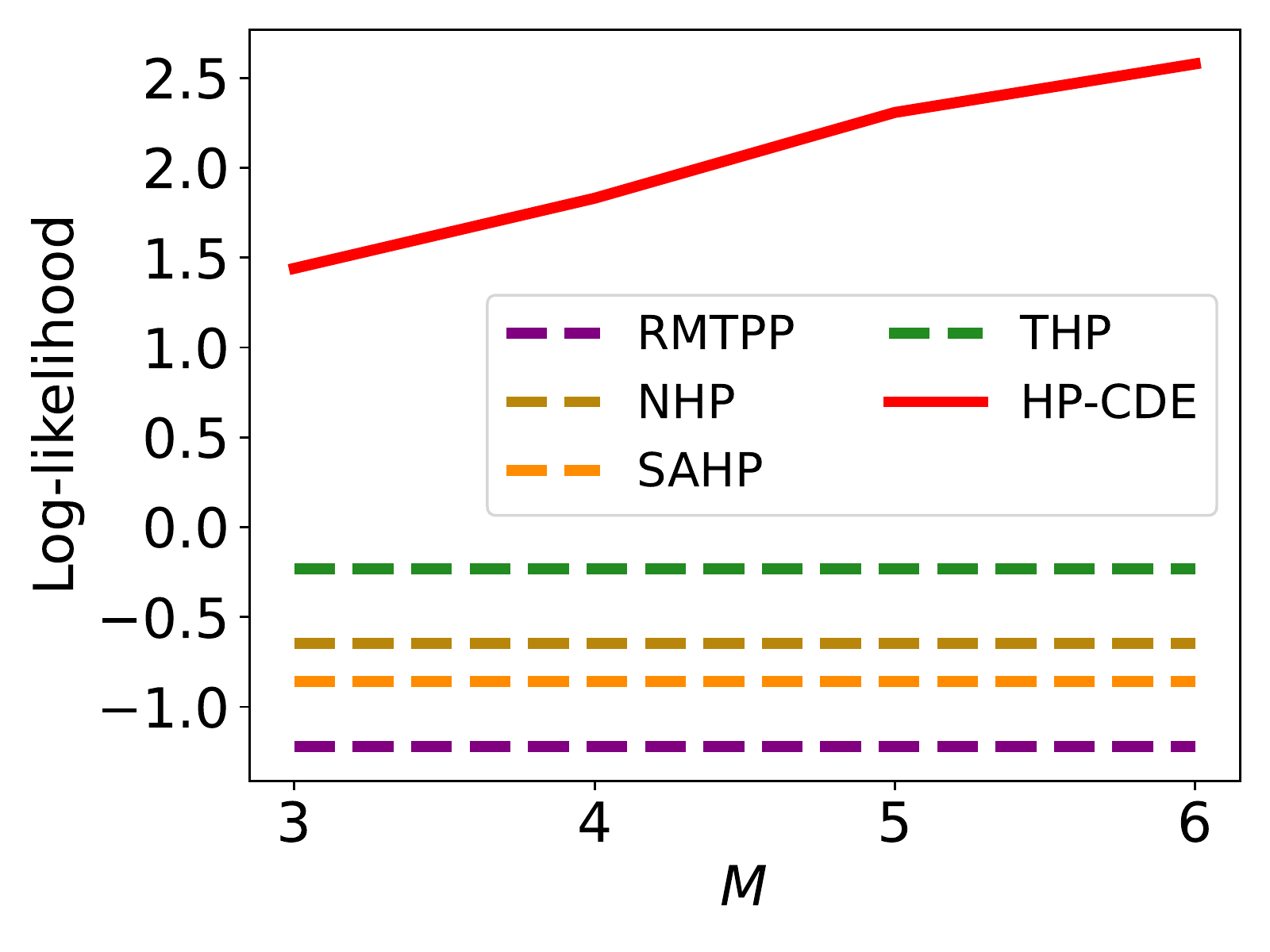}}
    \subfigure[Accuracy]{\includegraphics[width=0.48\columnwidth]{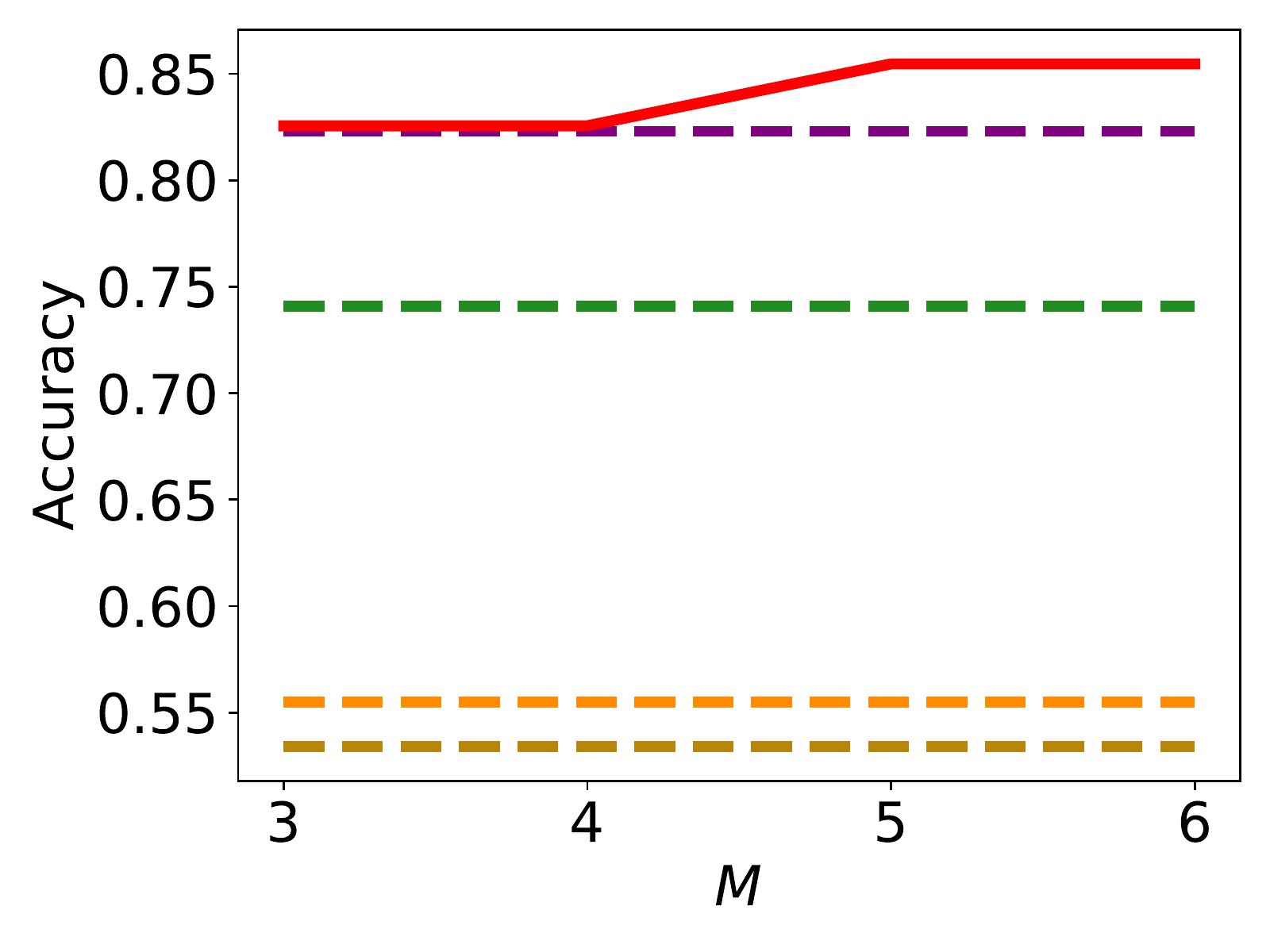}}
    \subfigure[RMSE]{\includegraphics[width=0.48\columnwidth]{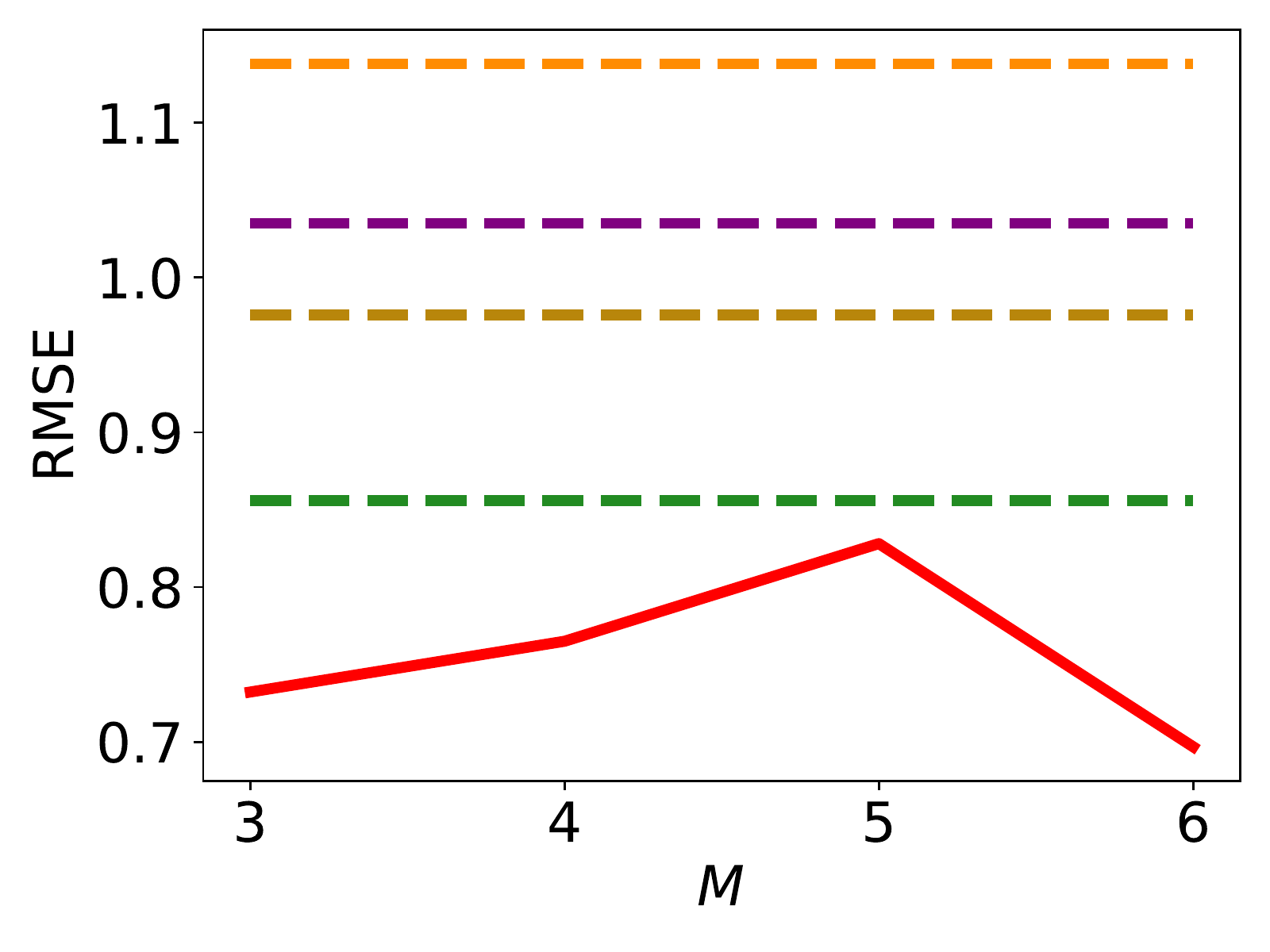}}
    \caption{Sensitivity study}
    
    \label{fig:sensitivity}
\end{figure}

We compare our model by varying the size of layers, $M$, in \{3, 4, 5, 6\}. As shown in Figure~\ref{fig:sensitivity}, HP-CDE consistently outperforms all the other baselines in every metric, regardless of $M$, presenting our model's robustness and efficacy. That is, continuous occurrence dynamics play a big role in modeling Hawkes process, by considering the intensity of events at every time point.

\section{Ablation Study}~\label{appendix:ablation}

As an ablation study, we apply a heuristic method to HP-CDE for evaluating the effectiveness of our proposed likelihood calculation method, which is using an ODE solver. We use the popular Monte Carlo method as a heuristic method since it is commonly used in baselines, and the fourth order Runge-Kutta method as an ODE solver. In Table~\ref{tbl:ablation}, we compare the results of the two methods when being applied to the likelihood computation of HP-CDE with all other conditions unchanged. According to Table~\ref{tbl:ablation}, our ODE solver-based HP-CDE shows better performance than that of the Monte Carlo method-based HP-CDE in both datasets. Moreover, the exact calculation of log-likelihood not only enhances the log-likelihood of sequences, but also improves the event prediction performance, in terms of accuracy and RMSE. Based on the results, it can be seen that our proposed likelihood computation method leads to better log-likelihood and prediction.



\section{Architectural Details of HP-CDE-AT }~\label{appendix:HP-CDE-AT}

As an ablation study on datasets with a long sequence length,  we put self-attention part of transformer right before the neural CDE layer in our proposed model. The additional layer consists of four stacked attention layers with two sub-layers: multi-head attention and feed-forward. In the multi-head attention sub-layer, we adopt the dot-product attention, which can be written as follows:
\begin{align}\label{eq:attention}
\begin{aligned}
\begin{split}
\text{Attention}(\mathcal{Q}, \mathcal{K}, \mathcal{V}) = \text{Softmax}\Big(\frac{\mathcal{Q}\mathcal{K}^{\top}}{\sqrt{\mathbf{dim(\mathcal{K})}}}\Big)\mathcal{V},\\
\text{where} \ \mathcal{Q} = E\mathbf{W_\mathcal{Q}}, \ \mathcal{K} = E\mathbf{W_\mathcal{K}}, \ \mathcal{V} = E\mathbf{W_\mathcal{V}},\\
\end{split}
\end{aligned}
\end{align}
where $E$ denotes to $\{\mathbf{E_e}(k_j) \oplus \mathbf{E_p}(t_j)\}_{j=1}^N$.
$\mathcal{Q}, \mathcal{K}$, and $\mathcal{V}$ are the query, key, and value matrices acquired from various transformations to $E$.
$\mathbf{W_\mathcal{Q}} , \mathbf{W_\mathcal{K}},
\mathbf{W_\mathcal{V}} \in  \mathbb{R}^{\dim(\mathbf{z}_j) \times \dim(\mathcal{K})} $ are the weights of $\mathcal{Q}, \mathcal{K}$, and $\mathcal{V}$ matrices. The feed-forward sub-layer consists of two linear transformations with a ReLU activation in between.

The output of the transformer layer $\{\mathbf{z}_j\}_{j=1}^N$ is equivalently matrix $\mathbf{Z} \in  \mathbb{R}^{N \times \dim(\mathbf{z}_j)}$ , where each row refers to a specific event. For example, $j$-th row of $\mathbf{Z}$ means the hidden state at time $t_j$ containing the information of j-th occurrence.

\section{Reproducibility} The implementation of our proposed method can be reproduced by the source code in our supplementary package. We will release our code for the sake of the public interest upon acceptance.


\end{document}